%% file: paper.tex
\documentclass{article}
\usepackage{spconf,amsmath,graphicx}
\usepackage{amssymb}
\graphicspath{{fig/}}

\title{Pattern Localization in Time Series through\\ Signal-To-Model Alignment in Latent Space}
\name{Steven Van Vaerenbergh$^1$, Ignacio Santamar\'ia$^1$, V\'ictor Elvira$^2$ and Matteo Salvatori$^3$ \thanks{The work of Steven Van Vaerenbergh was supported by the Ministerio de Econom{\'\i}a, Industria y Competitividad (MINECO) of Spain under grant TEC2014-57402-JIN (PRISMA).
The work of V\'ictor Elvira was supported by the Agence Nationale de la Recherche of France under PISCES project (ANR-17-CE40-0031-01).
}}

\address{$^1$  University of Cantabria, Santander, Spain.\\
$^2$ IMT Lille Douai CRISTAL (UMR 9189), Lille, France.\\
$^3$ Tecnatom S.A., Madrid, Spain.}

\begin{document}
%
\maketitle
\begin{abstract}
\input{abstract.txt}
\end{abstract}
\section{Introduction}
\label{sec:intro}

The localization of patterns in a time series is a common data mining problem, with applications in a multitude of fields, including biomedicine, finance, industrial engineering and speech recognition  \cite{fu2011review,esling2012time,ratanamahatana2004everything,liu2005locating}.
Contrary to the problem of detection, in which a decision is to be made about the presence or absence of a pattern, the problem of localization assumes that the pattern is present and its precise location is to be retrieved. %
The temporal nature of the data acquisition process complicates these tasks, as it causes the shape of the patterns of interest to suffer deformations in time known as \emph{warps}.
For pattern detection problems, many techniques exist based on aligning the query time series to a known reference pattern, commonly through dynamic time warping (DTW) \cite{berndt1994using}.
Similarly, a common pattern localization technique consists in aligning the query time series to a reference time series that contains several patterns of interest \cite{ratanamahatana2004everything}.

\begin{figure}
\includegraphics[width=\linewidth]{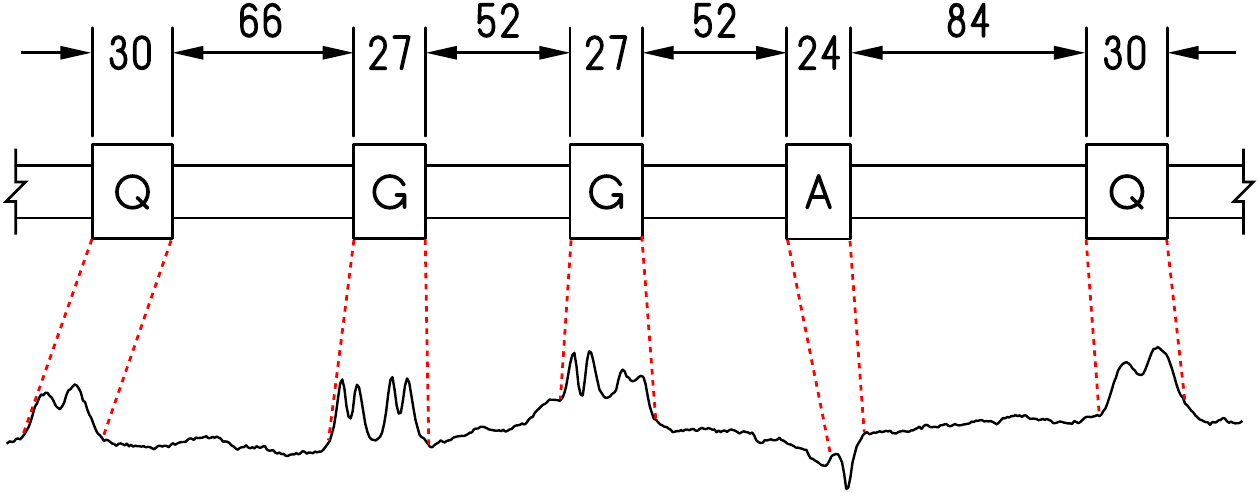}
\caption{Concept of the signal-to-model alignment problem. Top: A blueprint indicating the sizes (in mm) of references on a 1D body. Bottom: A time series acquired from the true physical structure. The dashed lines indicate the correct alignment between the model and the samples of the signal.}
\label{fig:concept}
\end{figure}

In many scenarios, however, the sequence of patterns to detect is not available as a reference time series but in the form of a blueprint that lists the theoretical locations of the expected patterns.
In this case, the localization problem translates to aligning the time series to a model (the blueprint).
A conceptual overview of this problem, which we refer to as ``signal-to-model'' (S2M) alignment, is given in Fig. \ref{fig:concept}.

The most prominent example of the S2M alignment problem in the scientific literature is ``audio-to-score'' alignment \cite{thickstun2017learning} in the context of music information retrieval.
There, a musical recording is to be aligned to the corresponding score.
S2M alignment is encountered in several other contexts as well, such as industrial environments, where a structure is measured in a one-dimensional fashion and the measured time series is to be aligned to a blueprint for this structure.
This is the scenario illustrated in Fig. \ref{fig:concept}, and a specific application in non-destructive testing will be discussed in detail in the experiments of Section \ref{sec:exp}.

The common approach to tackling the S2M alignment problem consists in first synthesizing a time series based on the available model, and then aligning the true time series to this artificial time series through standard alignment techniques for localization \cite{thickstun2017learning}.
The main bottleneck in this approach is the availability of an accurate synthesis technique.
In some contexts, such as music, techniques for synthesis are available that rely on pre-recorded waveforms and domain-specific knowledge.
In other contexts, the design of an appropriate synthesis technique often represents a complex problem.
An additional difficulty is due to variations in patterns that belong to the same class, illustrated by the pattern ``G'' in Fig. \ref{fig:concept}.
Even if an appropriate synthesis technique were available, the alignment would be hindered by these differences in the true waveforms.

In this paper, we propose a localization framework based on machine learning that increases the similarity between the synthesized and the true time series by mapping both into a latent \emph{correlation} space.
The mapping is learned from a set of training signals, and by assigning a high degree of freedom to this mapping, the choice of an appropriate synthesis technique becomes less critical.
We apply the proposed method to time series data acquired in non-destructive testing, obtaining a significant improvement over the state of the art.

\section{Signal-to-Model Alignment}
\label{sec:dtw}

\subsection{Problem definition}

We are given a time series $X = [x_1,x_2,\dots,x_N]$ and its corresponding theoretical model $\mathcal{M}$, consisting of a sequence of time markers $\mathcal{M} = \{(s_i,t_i,c_i)\}_{i=1,\dots,M}$.
Here, $s_i$ and $t_i$ mark the start and end locations of the $i$-th event of interest, respectively, and $c_i \in \mathcal{C}$ indicates the class of the event.
The start and end locations can be expressed in any unit that represents a one-dimensional quantity, for instance millimeters or seconds.
The localization problem can be solved by aligning the time series to the model $\mathcal{M}$, such that for each of the markers $s_i$ and $t_i$ the corresponding locations (expressed in samples) in the time series are retrieved.
We will denote the alignment solution as $\mathcal{A}$.

\subsection{Standard approach}

The standard approach to solving S2M alignment consists in transforming it into a setting of time-series alignment.
This, however, requires two time series, while the described scenario contains only one (in addition to the model). 
The second time series, $Y$, is synthesized from the information contained in the model. 
We briefly outline three synthesis strategies:

\begin{itemize}
\item \emph{Binary synthesis} generates a time series whose values are $1$ in the range $[s_i,t_i]$ of each event and $0$ outside. This is a rudimentary synthesis that may serve if the true patterns resemble rectangular blocks.
\item \emph{Replication synthesis} consists in replicating a true pattern at every position indicated by the model. This type of synthesis requires the availability of a \emph{training} signal from which patterns can be extracted.
\item \emph{Generative synthesis} encompasses advanced forms of synthesis that use domain-specific models to generate realistic time series. In music-alignment problems this type of synthesis is most often used \cite{thickstun2017learning}.
\end{itemize}

\begin{figure}
\includegraphics[height=2.7cm]{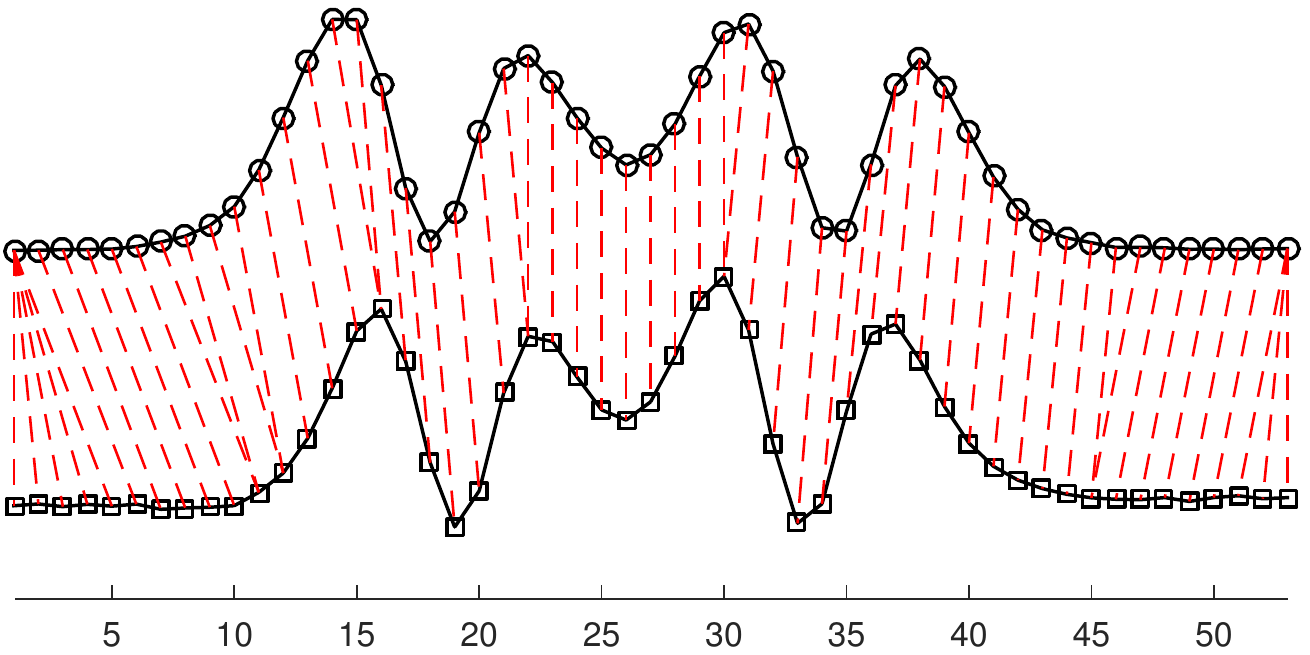} \hspace{\stretch{1}}
\includegraphics[height=3cm]{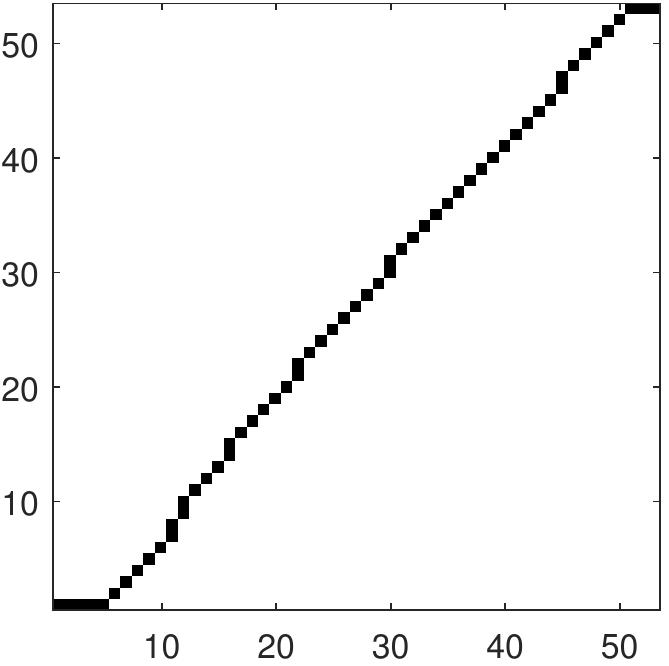}
\caption{Dynamic time warping. Left: Optimal alignment between two time series. Right: Corresponding warping path.}
\label{fig:dtw}
\end{figure}

Once the synthesized time series $Y$ is available, it is aligned to the true time series $X$, typically through dynamic time warping \cite{rabiner1993fundamentals,vintsyuk1968speech}.
DTW is a well-known technique and the de-facto standard algorithm for aligning time series.
By relying on dynamic programming, DTW allows to evaluate a combinatorially large number of warps, each of which consists of local shifts, contractions, and stretches of the signals. 
Figure \ref{fig:dtw} illustrates an alignment solution obtained by DTW, and its corresponding warping path.
The alignment solution, denoted as $\mathcal{A}$, contains the warping path, which consists of the pairs of samples from $X$ and $Y$ that are aligned.

The solution $\mathcal{A}$ allows to map any location in the model $\mathcal{M}$ (across $Y$) to a sample index in the true time series $X$, and vice-versa, hence solving the localization problem.

\subsection{Challenges}

As mentioned in the introduction, the main bottleneck in the standard approach lies in the availability of an accurate synthesis technique.
In problems with little domain-specific knowledge, for instance, only basic forms of synthesis can be applied, which may not guarantee sufficient similarity between the synthesized and the true time series.
Furthermore, even if a realistic synthesis procedure were available, the patterns in the true time series may show variations that hinder the final alignment.

In order to mitigate the differences between the synthesized and the true time series, an additional transformation is sometimes performed on both.
For music, a chromagram can be used, which is a condensed form of the spectral information, representing notes \cite{thickstun2017learning}.
Nevertheless, such transformations are not guaranteed to be optimal, and in essence they require the availability of domain-specific information.


\begin{figure*}
\centering
\includegraphics[height=2.7cm]{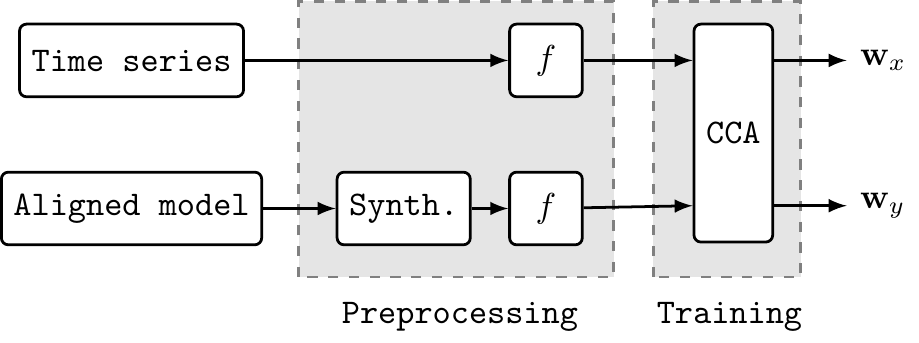}
\hspace{\stretch{1}}
\includegraphics[height=2.7cm]{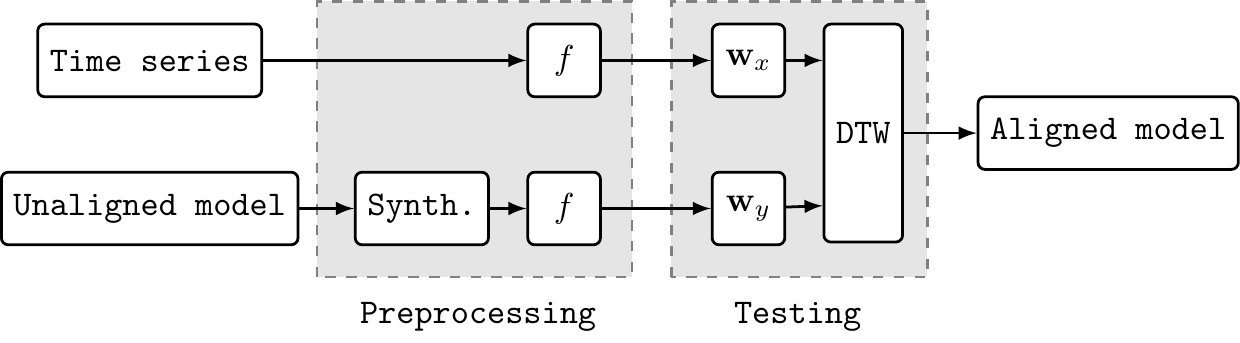}
\caption{Overview of the proposed machine learning framework. Left: Training setup. The block ``Synth.'' represents the synthesis of a time series given a model $\mathcal{M}$, and $f$ represents a predefined transformation. Right: Test setup. After preprocessing, testing consists of mapping the time series to the latent space through the projections $\mathbf{w}_x$ and $\mathbf{w}_y$, and finally DTW is performed.}
\label{fig:diagrams}
\end{figure*}

\section{Proposed technique}
\label{sec:cca}

We propose to automatically learn a transformation that maximizes the similarity between the synthesized and the true time series.
This transformation can be interpreted as a mapping to a latent space where both time series are more similar.
The goal of the mapping, therefore, is to emphasize the common components in both time series, and to suppress noise signals and artifacts present in only one of them.

The learning process is illustrated in the diagram of Fig. \ref{fig:diagrams}, left plot.
In order to learn the mappings, a \emph{training} time series $X$ is required, together with a model that has been aligned to $X$, for instance by a human labeler.
A synthetic time series $Y$ is first obtained from the aligned model, which now represents a time series that is perfectly aligned to $X$.

Next, the optimal transformations need to be learned.
The transformations used in this work are linear filters, which are implemented as projections of time-embedded vectors.
The transformations $f$ in the diagram, therefore, correspond to time embeddings in this case. In order to obtain the optimal mappings, the technique of canonical correlation analysis is used, which we discuss in the sequel.

%
%
%
%

\subsection{Learning the latent correlation space}

Given two multidimensional random variables $\mathbf{x}$ and $\mathbf{y}$, canonical correlation analysis (CCA) seeks a pair of optimal linear transformations such that the transformed variables are maximally correlated \cite{hotelling1936relations,hardoon2004canonical}.
In the present context, $\mathbf{x} \in \mathbb{R}^{M}$ and $\mathbf{y} \in \mathbb{R}^{M}$ represent samples from the time series that have been time-embedded with an embedding of size $M$.
Denote by $\mathbf{w}_x$ and $\mathbf{w}_y$ the respective projection vectors, and by $x^l = \mathbf{x}^\top \mathbf{w}_x$ and $y^l = \mathbf{y}^\top \mathbf{w}_y$ the samples from the obtained transformed time series in the latent space, $X^l$ and $Y^l$.
If we represent the covariance matrix between $\mathbf{x}$ and $\mathbf{y}$ by $\mathbf{R}_{xy}$, and equivalently the autocovariance matrices $\mathbf{R}_{xx}$ and $\mathbf{R}_{yy}$, the function to be maximized is
\begin{equation}
\rho = \max_{\mathbf{w}_x,\mathbf{w}_y} \frac{\mathbf{w}_x^\top \mathbf{R}_{xy} \mathbf{w}_y}{\sqrt{\mathbf{w}_x^\top \mathbf{R}_{xx} \mathbf{w}_x \mathbf{w}_y^\top \mathbf{R}_{yy} \mathbf{w}_y}}.
\end{equation}
This problem can be solved as a generalized eigenvalue problem, which yields the optimal projectors $\mathbf{w}_x$ and $\mathbf{w}_y$, see \cite{hardoon2004canonical} for additional details.
Note that the training process is not limited to a single time series, as the covariance matrices can be easily constructed for an entire set of training data.

\subsection{Testing on new time series}

When a new time series is considered, alignment can now be performed in a straightforward fashion as illustrated by the diagram of Fig. \ref{fig:diagrams}, right plot.
Apart from the time series, the proposed technique requires the corresponding model as an input, which is not aligned yet at this point.
Similarly to the training process, the model is synthesized into a time series, and a fixed transformation $f$ (the  embedding) is applied to both time series.
Then, the mapping to the latent space is performed, by projecting the true time series with $\mathbf{w}_x$ and the synthesized series with $\mathbf{w}_y$.
The obtained time series in the latent space, $X^l$ and $Y^l$, now show a high similarity and can be aligned by applying DTW.
The alignment solution $\mathcal{A}$ is finally used to produce a model that has been aligned to the true time series (as shown in the diagram), or, equivalently, an alignment of the true time series to the original model.

The entire transformation of the time series consists in the fixed transformation $f$ followed by the projection to the latent space.
If $f$ represents a large time embedding, $M\gg 1$, the projections possess a large number of degrees of freedom to define an optimal one-dimensional latent space.
In some preliminary experiments, we have verified that this flexibility may compensate for poor forms of synthesis, such as the binary synthesis mentioned earlier.
As a result, the proposed machine-learning based technique represents a framework that can be easily applied to S2M alignment problems in which domain-specific knowledge is scarce.


\section{Numerical Experiments}
\label{sec:exp}

We apply the proposed technique to time series acquired in non-destructive testing of heat generator tubes \cite{garcia2011non}.
The data consists of $198$ time series acquired through eddy current testing, and for each time series a corresponding blueprint is available that indicates the theoretical locations of support structures, similar to the model represented in Fig. \ref{fig:concept}.
Each support structure leaves a characteristic pattern in the measured time series.
Due to space restrictions, we will only consider a simplified scenario with one class of support patterns.

\begin{figure}
\includegraphics[width=\linewidth]{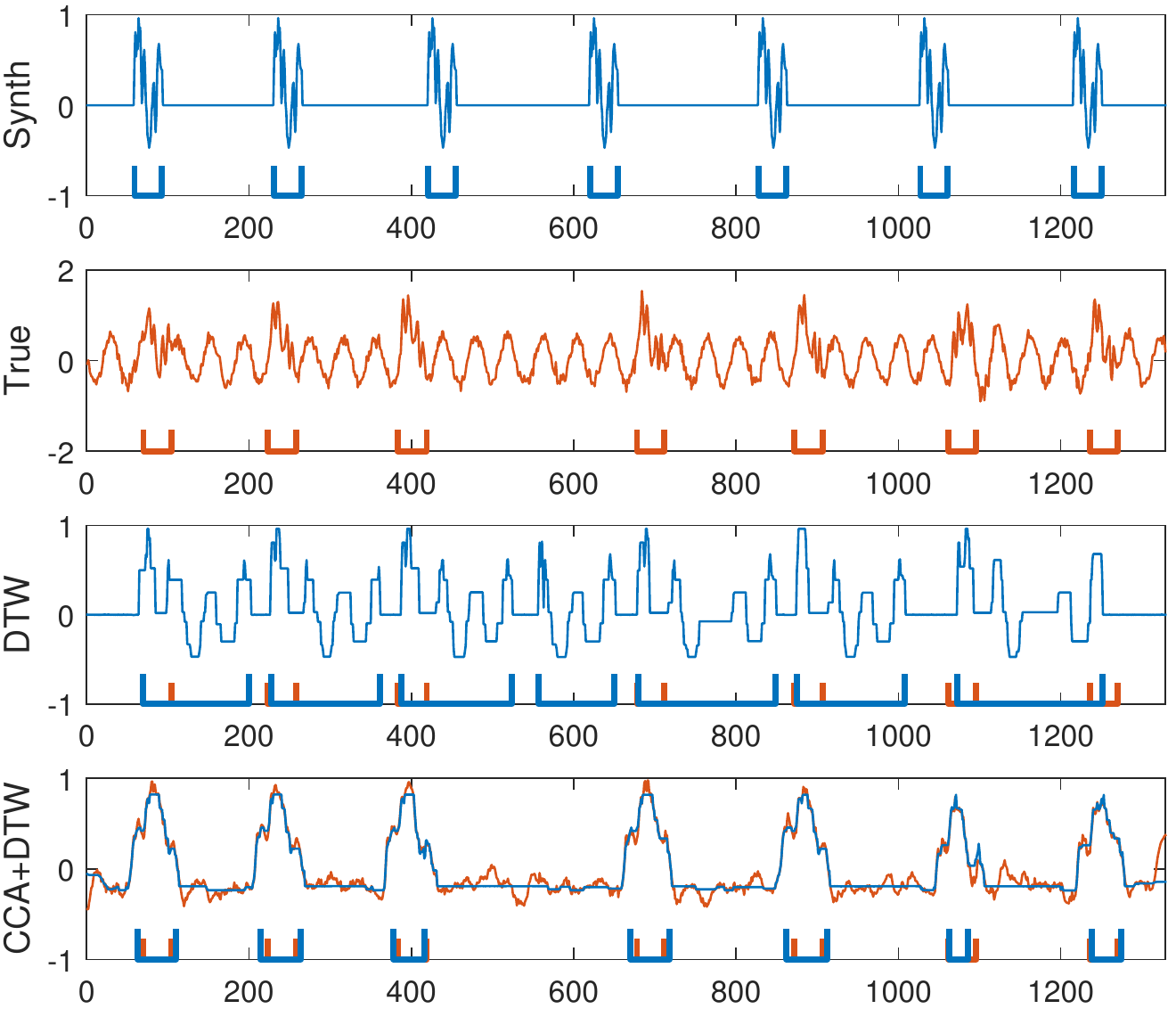}
\caption{Results for the first experiment. Top: Synthesized time series, with pattern starts and ends marked below. Second plot: True time series, with true starts and ends marked. Third plot: Result of aligning Synth to True using DTW; markers are shown as estimated by the alignment as well as the true markers. Bottom: Alignment of the proposed CCA+DTW technique; The latent representations for both Synth (blue) and True (red) are shown, and their markers.}
\label{fig:exp1}
\end{figure}

First, we study the basic alignment problem, in which one training time series is used to extract a pattern for replication synthesis, and a second time series is used for testing the localization technique.
To both time series we artificially add a common background noise known as ``pilgrim noise'' \cite{benoist1991expert}, and we further add a small temporal warp.
The synthesized and the test time series are shown in Fig. \ref{fig:exp1}, first two plots.

As a benchmark, we apply the DTW algorithm to align both series directly.
The third plot of Fig. \ref{fig:exp1} shows the synthesized time series after applying the found time warp.
Clearly, the warping path does not correspond to a valid alignment.
For instance, DTW determined that there is a pattern that includes sample $600$ while there is none, which causes the localization of all consecutive patterns to shift.
On average, the error between the true markers provided by a human labeler and markers estimated through time warping of the model amounts to $85.93$ samples.

Next, we apply the proposed method with an embedding of $20$ samples from the past and $20$ from the future, i.e. $M=41$.
Both time series are first mapped to a latent space that is learned from the training time series, after which standard DTW is applied.
The last plot of Fig. \ref{fig:exp1} shows the projections, in the latent space, of the true series and the aligned synthesized series.
The estimated markers, shown below together with the true markers, have an average error of $5.86$ samples.

\begin{figure}
\includegraphics[width=\linewidth]{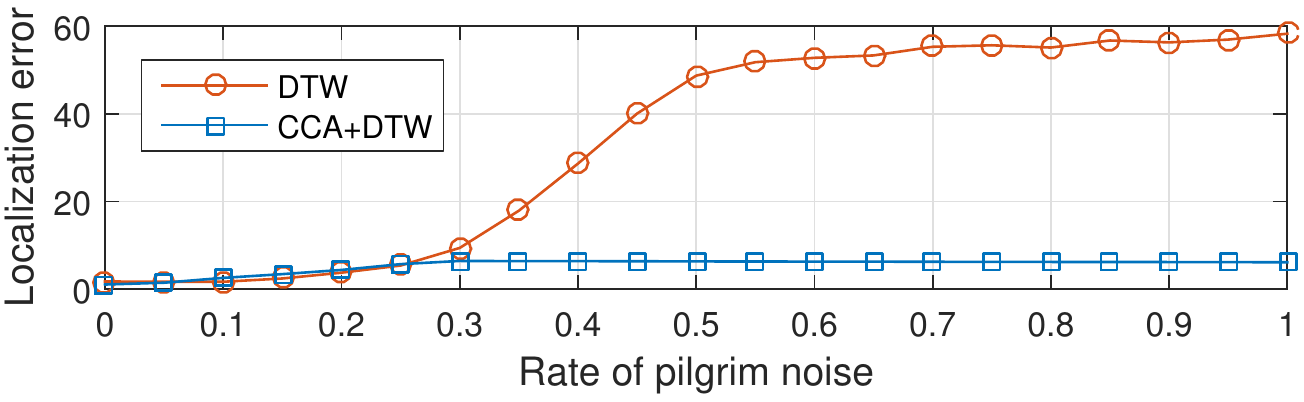}
\caption{Mean localization errors obtained by both methods in the second experiment, for different rates of sine noise.}
\label{fig:exp2}
\end{figure}

In the second experiment, we test the alignment techniques for different degrees of pilgrim noise, ranging from $0$ (no noise) to $1$ (noise of the same amplitude as the maximum peak in the signal).
A single pattern was used for creating the synthesis time series, and for learning the latent space a set of $19$ training time series was used.
The remaining time series were used for testing.
The optimal embedding for each amount of noise was determined by cross-validation on the training data.

Fig. \ref{fig:exp2} shows the average localization error for DTW and for the proposed CCA+DTW technique.
For low rates of noise, both techniques perform similarly.
Starting from a noise rate of $0.3$, however, the benefit of operating in the learned latent space becomes clear.

\section{Conclusions and discussion}
\label{sec:concl}

We have presented a novel technique for locating patterns in time series for which a blueprint model is available.
Similar to other methods in the literature, the proposed method operates by synthesizing the model into a time series, in order to perform dynamic time warping and aligning both series.
In addition, though, the proposed technique maximizes the similarity between both time series, before performing DTW, by mapping them to a latent space where the time series are maximally correlated.
This optimal mapping is learned through canonical correlation analysis.
Experiments with eddy current testing data show that the proposed technique is capable of correctly locating patterns in time series in challenging scenarios with high degrees of background noise.

The design of the method does not rely on any domain-specific knowledge, and is therefore expected to operate satisfactorily in a wide range of applications.
In future research, we plan to apply the proposed method among others to the audio-to-score alignment problem.
Due to space restrictions, we have limited the experiments in this paper to a simplified setup that highlights the strength of the proposed method.
Nevertheless, preliminary experiments with more complex setups, including patterns of several different classes, multidimensional time series, and additional noise types, have been successful and will also be the topic of future research.

\section{Acknowledgments}
The authors thank Tecnatom S.A., Madrid, Spain, for providing the data used in the experiments.


\bibliographystyle{IEEEbib}
\bibliography{biblio}

\end{document}

%% file: abstract.txt
In this paper, we study the problem of locating a predefined sequence of patterns in a time series. In particular, the studied scenario assumes a theoretical model is available that contains the expected locations of the patterns. This problem is found in several contexts, and it is commonly solved by first synthesizing a time series from the model, and then aligning it to the true time series through dynamic time warping. We propose a technique that increases the similarity of both time series before aligning them, by mapping them into a latent correlation space. The mapping is learned from the data through a machine-learning setup. Experiments on data from non-destructive testing demonstrate that the proposed approach shows significant improvements over the state of the art.